\def\ANON{0} 

\documentclass[conference]{IEEEtran}
\usepackage{cite}
\usepackage[dvips]{graphicx}
\graphicspath{{./eps/}}
\DeclareGraphicsExtensions{.eps}
\usepackage{amsmath,amssymb,color}
\usepackage{amsmath}
\usepackage{algorithm}
\usepackage{algpseudocode}
\usepackage{booktabs} 
\usepackage[table,xcdraw]{xcolor}
\usepackage{array}
\usepackage{mdwmath}
\usepackage{mdwtab}
\usepackage{eqparbox}
\usepackage{multirow}
\usepackage{float}
\usepackage{verbatim}
\usepackage{graphicx}
\usepackage{subcaption}
\usepackage{fixltx2e}
\usepackage{url}
\usepackage[hidelinks]{hyperref} 
\usepackage{balance}
\usepackage{flushend}
\IEEEoverridecommandlockouts

\usepackage{url}

\usepackage{xspace}
\usepackage{xcolor}
\definecolor{orange9}{HTML}{FFDD00}
\usepackage[color=orange9]{todonotes}

\begin{document}
	\title
     {Analyzing the vulnerabilities in SplitFed Learning: Assessing the robustness against Data Poisoning Attacks}

\ifnum\ANON=1
    \author{
    \IEEEauthorblockN{Anon}
    \IEEEauthorblockA{\\ \\}
    \email{}
    }
\else
    \author{
     \IEEEauthorblockN{Aysha Thahsin Zahir Ismail and Raj Mani Shukla }
     \IEEEauthorblockA{Computing and Information Science,\\  Anglia Ruskin University, Cambridge, UK}
     \IEEEauthorblockA{
    \tt \{az303, raj.shukla\}@aru.ac.uk}
    }
\fi

\maketitle

\textbf{\begin{abstract}
Distributed Collaborative Machine Learning (DCML) is a potential alternative to address the privacy concerns associated with centralized machine learning. The Split learning (SL) and Federated Learning (FL) are the two effective learning approaches in DCML. Recently there have been an increased interest on the hybrid of FL and SL known as the SplitFed Learning (SFL). This research is the earliest attempt to study, analyze and present the impact of data poisoning attacks in SFL.  We propose  three kinds of novel attack strategies namely untargeted, targeted and distance-based attacks for SFL. All the attacks strategies aim to degrade the performance of the DCML-based classifier. We test the proposed attack strategies for two different case studies on Electrocardiogram signal classification and automatic handwritten digit recognition. A series of attack experiments were conducted by varying the percentage of malicious clients and the choice of the model split layer between the clients and the server. The results after the comprehensive analysis of attack strategies clearly convey that untargeted and distance-based poisoning attacks have greater impacts in evading the classifier outcomes compared to targeted attacks in~SFL.  
\end{abstract}
}

\section{Introduction}
\label{introduction}
Artificial Intelligence (AI) and Machine learning (ML) are being deployed by a wide range of organizations worldwide, from governments and massive tech companies to small internet retailers. 83\% of the tech industry utilizes AI-powered technologies for developing applications \cite{howarth_2023}. With significant improvement in productivity and performance, ML can impart efficiency in several domains such as product recommendation, biomedical image classification, computer vision, and natural language processing \cite{shetty2022supervised}. 

Most ML applications employ supervised ML model \cite{shetty2022supervised}. The performance of ML models in actual application scenarios  depends on the quality of training data. To achieve improved model performance and accuracy, machine learning systems require a huge amount of quality training samples which might be split among various groups \cite{hu2019ACM}. In addition, it is often difficult to obtain labelled training samples\cite{bansal2022acm}. Further, gathering all the training samples to a centralized server has several privacy concerns, especially during the presence of sensitive information. Several privacy governing regulations such as the General Data Protection Regulation (GDPR) must be complied with while aggregating private data into a central server. Distributed Collaborative Machine Learning (DCML) is a potential alternative that enables multiple participants to collaboratively train a shared global model while locally keeping their training data. This technique allows the participants to share the updates with the global model analysts without any access to local training data \cite{guo2021arxiv}. 

Federated learning \cite{li2020ieee} and Split learning \cite{vepakomma2018arxiv} are DCML approaches that resolve the privacy issues in centralized ML. In Federated Learning (FL), multiple clients train an entire machine learning model with their local training samples and further, the locally trained models of all clients are aggregated to obtain a global model at the server. Though FL prevents sharing of local data, it is not viable when clients have limited resources for computing large ML models. In addition, both the server and clients can access local and global models affecting the privacy of clients training data and model parameters of the server. Additionally, communication delays, the presence of heterogeneous systems in distributed learning, and data dynamism are other challenges experienced by FL with multiple clients. Split learning (SL) was introduced to overcome these issues such that resource constraints and model privacy by splitting the ML model between the client and the server. SL ensures that the client and server will have access to a portion of their split of the whole ML model \cite{gupta2018Elsevier}. However, SL is not ideal in the presence of many clients as it can train only one client at an instance which eventually idles other clients and leads to longer training time \cite{thapa2022AAAI}.

SplitFed Learning (SFL) is an advanced DCML paradigm that resolves the issues caused by FL and SL. SFL has a hybrid architecture where the model is split as in SL which overcomes limited client resources followed by parallel computation as in FL to mitigate the training overhead that occurs during the presence of a large number of clients \cite{thapa2022AAAI}. Various studies have evaluated the security of FL. FL is prone to model poisoning attacks that manipulate gradients to minimize accuracy \cite{fang2020USENIX}; \cite{shejwalkar2022ieee}. Inference attacks, which attempt to recreate private data from the client or server, have a substantial impact on the security of SL \cite{pasquini2021ACM}.   
However, there exists minimal research analyzing the susceptibility of SFL against adversarial attacks where training data is spread among multiple clients.

This work examines how a malicious client can initiate data poisoning attacks in the SFL system. Data poisoning attacks attempt to manipulate training data that eventually influences the learning output of the trained model. Data poisoning attacks broadly take the form of clean label poisoning and dirty label poisoning where the former injects tampered data into the train set and the latter manipulates the training labels such as label flipping \cite{lin2021arxiv}. 

Accordingly, the major contribution of this paper can be summarized as follows: 

\begin{enumerate}

\item This research proposes targeted, untargeted and distance-based data poisoning attacks on SFL to evade the aggregated model outcomes. 
 
\item The research tests the proposed targeted, untargeted, and distance-based data poisoning attack strategy for two case studies – 1) hand-written digit classification (standard MNIST dataset) and 2) a novel application of ECG signal classification for arrhythmia heartbeat type using SFL.  

\item This paper conducted an extensive study on the proposed attacking strategies on SFL varying the proportion of model split and malicious client percentage in MNIST and health care ECG signal datasets.

\end{enumerate}

To the best of our knowledge, this paper is the first attempt --  i) to employ privacy-preserving SFL for the automated classification of ECG signals, ii) in attacking SFL using  targeted and untargeted data poisoning strategies for the proposed ECG classification problem iii) in assessing SFL's sensitivity to a novel distance-based attacks.

The rest of this paper is organized as follows: Section~\ref{section:background} presents a comprehensive discussion regarding the existing literature. Section~\ref{background} provides a quick overview of the different DCML techniques. Section~\ref{section:methodology} presents the proposed attack techniques in SplitFed Learning.  Section~\ref{cha:implementation} discusses the implementation specifics, including the system architecture, datasets, and set-up for a poisoning attack, while section~\ref{cha:results} outlines the results and the  performance of the poisoning attacks with respect to two case studies. Finally, the section~\ref{conclusions} concludes this paper. 

\section{Related Work} 
\label{section:background}

Numerous adversarial attacks are experienced by federated learning mainly poisoning attacks and information extractions \cite{sun2022ieee}. Model poisoning and data poisoning attacks are the major security threats encountered by FL. In Data poisoning attacks, the attacker introduces malicious data samples into training data changing their primary meaning before the training phase leading to incorrect results. In contrast, Model poisoning attacks manipulate the machine learning model rather than the data changing overall learning outcomes \cite{duan2022sensors}. Most of the adversarial backdoor attacks in FL manage to manipulate the local client update or the training data among the edge devices \cite{lin2020ACM}.  Lyu, et al. (2020) classified malicious actors in FL into 3, namely malicious server, insider, and outsider adversary. The impact of poisoning attacks by untrusted participants critically damaged the performance of FL \cite{lyu2020arxiv}. Besides the poisoning attacks, the security of FL is challenged by inference attacks induced by dishonest or malicious servers. These servers are capable of learning and extracting clients’ private data using their gradient updates \cite{lee2021IEEE}.  

Similar to FL, there are several privacy threats to SL mainly due to training data inference from the intermediate representation generated by smashed data, label leakage of client data, and client model inversion \cite{duan2022sensors}. Pasquini et al. (2021) implemented a Feature-Space Hijack Attack (FSHA) on the SL model in which an untrusted server retrieves the private data of a client that is used to train the model. The hijack occurs in two phases: the setup phase, during which the server seizes the client training process, and the inference phase, during which the server obtains the client training data using the smashed data received from the client. 
Erdoğan, et al.  studied the possibility of model stealing in SL and formulated a stealing attack that can cause client model inversion \cite{erdougan2022unsplit}. In a Two-Party Split Learning, there will be a data owner (client) and label owner (server), and private label leakage attacks take place in this setting when any external adversary or clients attempt to infer the private labels \cite{duan2022sensors}. Li, et al.  presented a label leakage attack by analysing the gradient norm of imbalance classes present in the training set \cite{li2021arxiv}. Several distance correlations and differential privacy strategies were implemented to improve the security in SL \cite{vepakomma2020ieee}; \cite{titcombe2021arxiv}. Hence, combining FL and SL may fully leverage the strengths of both learning approaches while minimising their individual limitations.

SFL is a hybrid DCML architecture that is a combination of FL and SL where it combines the parallel training/testing of client-side models as observed in FL and the model split between client and server as performed in SL. The SFL systems consist of client and server segments along with an additional server called the fed server on the client side. The fed server is used to perform FedAvg aggregation algorithm on the updates provided by the client and is responsible for synchronising the global model updates of multiple clients \cite{thapa2022AAAI}.  
Each client in SFL parallelly performs forward propagation on the client-side model split with their local training data until the cut layer, server then proceeds with the forward and backward propagation as in SL and sends the updated gradients to all of its clients in parallel. Further, each client completes the backward pass on their client-side model, and updates are forwarded to the fed server. Fed server conducts FedAvg on the updates from all clients resulting in a client-side global model and the parameters redirected to all clients \cite{thapa2021springer}.
SFL effectively addresses the difficulties that FL and SL encounter and provides greater privacy than FL. Yet, there is a major possibility of data poisoning attacks while collaboratively training among distributed clients and a server with SFL. Malicious participants can induce poisonous data during the training process that is difficult to detect by the aggregator. 

Motivated by the aforementioned analysis, this research introduces data poisoning attacks against SFL and tries to fill the research gap that studies the robustness of SFL. This paper proposes targeted, untargeted along with a novel distance-based attack strategy and performs a comparative analysis of the proposed attacks on  MNIST and healthcare ECG signal datasets. Our work is the closest to \cite{GajbhiyeRTIMPR2022}. However, in contrast to \cite{GajbhiyeRTIMPR2022}, we analyzed the robustness of splitFed learning for the novel ECG classification problem. We also proposed a new distance-based attack technique in our work in contrast to \cite{GajbhiyeRTIMPR2022}.

\vspace{-2cm}
\section{Background}
\vspace{-1.8cm}

\label{background}
This section provides a basic background of Federated Learning (FL), Split Learning (SL), and SplitFed Learning (SFL). 

\subsubsection{Federated Learning} 
\begin{figure}[H]
\hspace*{-2cm}                                                           
\includegraphics[width=1.5\linewidth]{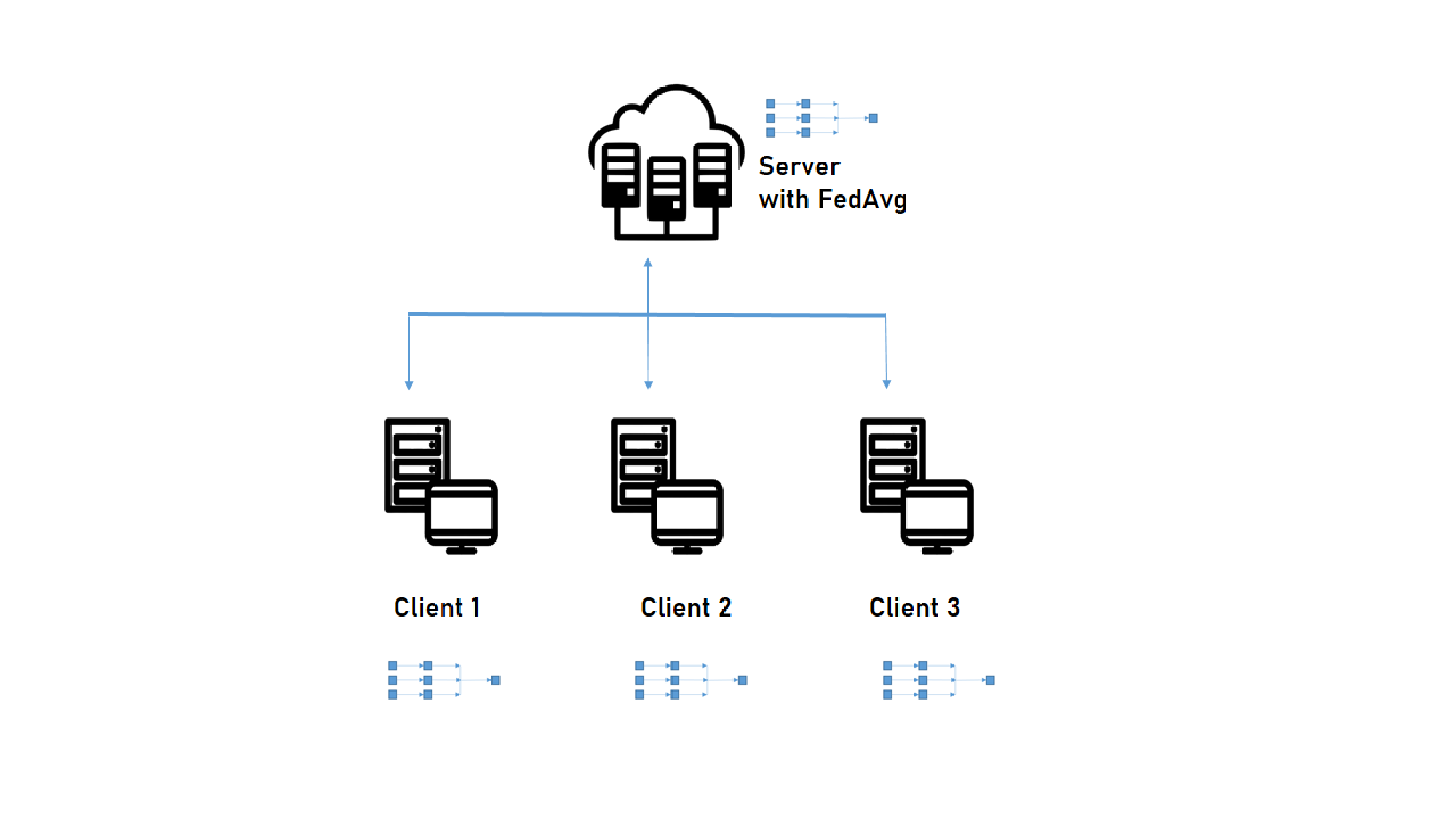}
\centering
\caption{Federated Learning }  
\label{fig:federated_learning}
\end{figure}

The FL is pictorially represented in Figure ~\ref{fig:federated_learning}. The fundamental idea behind FL is collaborative training of ML models among distributed data holders. In a decentralized setting with multiple clients, each client has its local data and trains the complete ML model. After each training iteration, all clients transfer the updated weights obtained while computing forward and backward passes on its local model to a central server. FedAvg, is a commonly used aggregation algorithm that is employed by the server to achieve a global update for the ML model. This global update is passed on to the clients for the subsequent iteration. Instead of sharing raw client data for training as is seen in traditional centralized ML, FL only shares the parameters of the model with the server or other clients. Following that, FL lowers communication costs and eases the networking overhead involved in Internet of Things (IoT) services with various entities and limited resources \cite{mcmahan2017ais}.

\begin{figure}[H]
\hspace*{-0.4cm}                                                           
\includegraphics[width=1.2\linewidth]{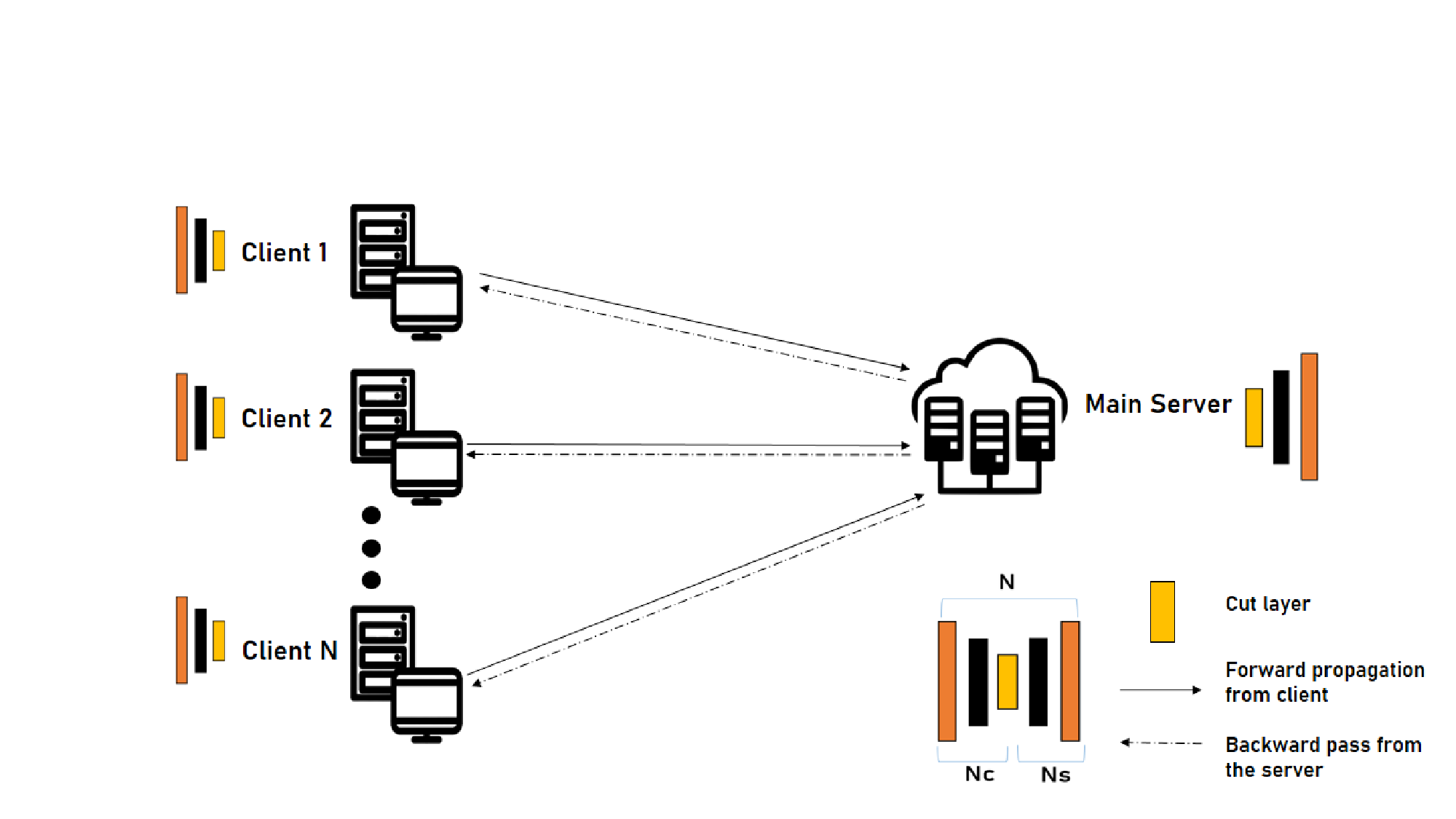}
\centering
\caption{Split Learning }  
\label{fig:split_learning}
\end{figure}

\begin{figure}[H]
\hspace*{-0.4cm}                                                           
\includegraphics[width=1.2\linewidth]{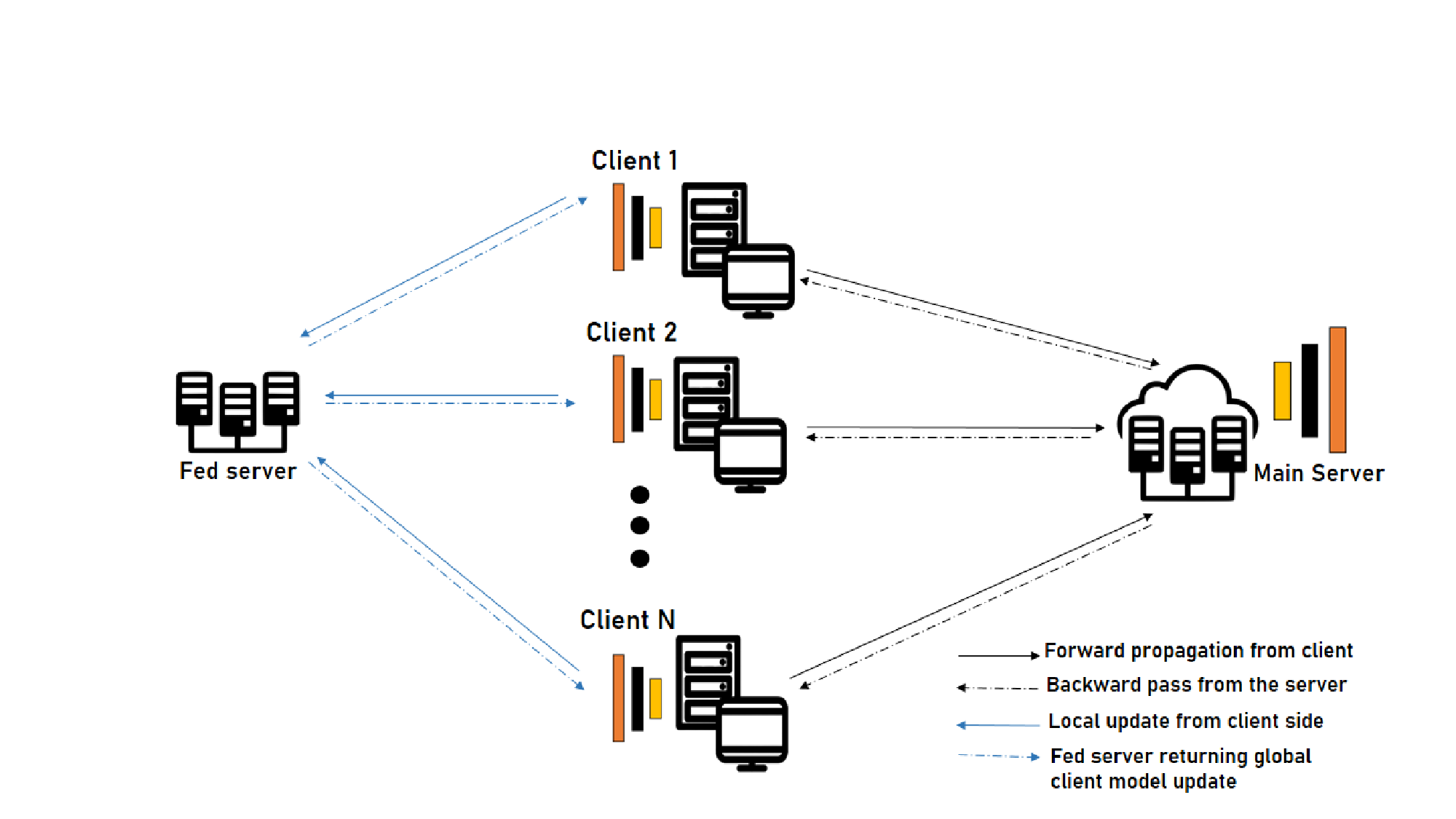}
\centering
\caption{SplitFed Learning}  
\label{fig:splitfed_learning}
\end{figure}

\subsubsection{Split Learning} Figure~\ref{fig:split_learning} presents the basic architecture of SL. SL \cite{gupta2018Elsevier} is a DCML approach that divides the ML/DL model between the client and server. The model layer at which the split occurs is referred to as the cut layer and the output generated is termed smashed data. Computations on initial model layers are performed by the client and the later layers are handled by the server, thereby local training data is kept private similar to FL. In SL sequential training is performed where each client performs forward propagation with its own model segment until the \emph{cut layer} which is the last layer of the client-side model split. The \emph{smashed data} (activations from the cut layer) from the cut layer is received by the server which continues to propagate forward with the model’s remaining layers \cite{thapa2021springer}.   Once the forward propagation is completed, the server determines the loss and begins backpropagation. The gradient calculated up to the cut layer is passed on to the client to continue its backpropagation. This entire process is one training round and the updates are sent to the next client \cite{thapa2021springer}.

\subsubsection{SplitFed Learning}
The basic architecture of SFL is represented in the Figure~\ref{fig:splitfed_learning}. SFL-based distributed client environments include a main server, a fed server, and a group of clients as represented in the Figure. The full model $N$ is split into client-side model $N^C$ and server-side model $N^S$. At each global epoch, all clients interact with the server in parallel and the main server aggregates the parameters to generate a global server-side model. The client model synchronization is carried out in parallel at the fed server. Considering \emph{k} clients at time instance \emph{t}, the client-side model of each client can be represented as  $N_{k,t}^{C}$. The smashed data of each client at \emph{t} is $S_{k,t}$.  At \emph{t}=0, each client \emph{k} performs forward propagation of its model split and sends the activations  $S_{k,t}$  along with the true labels to the server \cite{thapa2022AAAI}.

The server on receiving the smashed data carries out forward propagation with its model split, computing the prediction, and loss calculation with actual labels and predicted labels $\hat{y}$. Further, the server executes the global server-side model update and propagates the gradient back to the client. Simultaneously, when each client receives the back propagated smashed data from the server, it is sent to the fed server to aggregate and generate a global client-side model update that is sent back to all \emph{k} the clients \cite{thapa2022AAAI}. Table~\ref{tab:Notations} presnets the basic notations of the SFL.


\begin{table}[hbt!]
\centering
\caption{Principle Notations of SFL}
\label{tab:Notations}

\begin{tabular}{@{}ll@{}}
\toprule
\textbf{Symbol}     & \textbf{Definition} \\ \midrule
\textit{N}          & The full deep learning model                    \\ 
$N^C$               & Client-side model split                         \\ 
$N^S$               & Server-side model split                         \\ 
$S_{k,t}$           & Smashed data for client k at time t             \\ 
$y$,$\hat{y}$       & Actual labels and predicted labels respectively \\ \bottomrule
\end{tabular}

\end{table}

\section{Methodology: Data Poisoning Attacks in SplitFed Learning } 
\label{section:methodology}
This section discusses the proposed methodology of data poisoning attacks on SFL. We discuss the threat model and  algorithms used to attack SFL-based DCML.


\subsection{Proposed threat Model in SplitFed System}
\begin{figure*}[hbt!]
\includegraphics[width=1\textwidth]{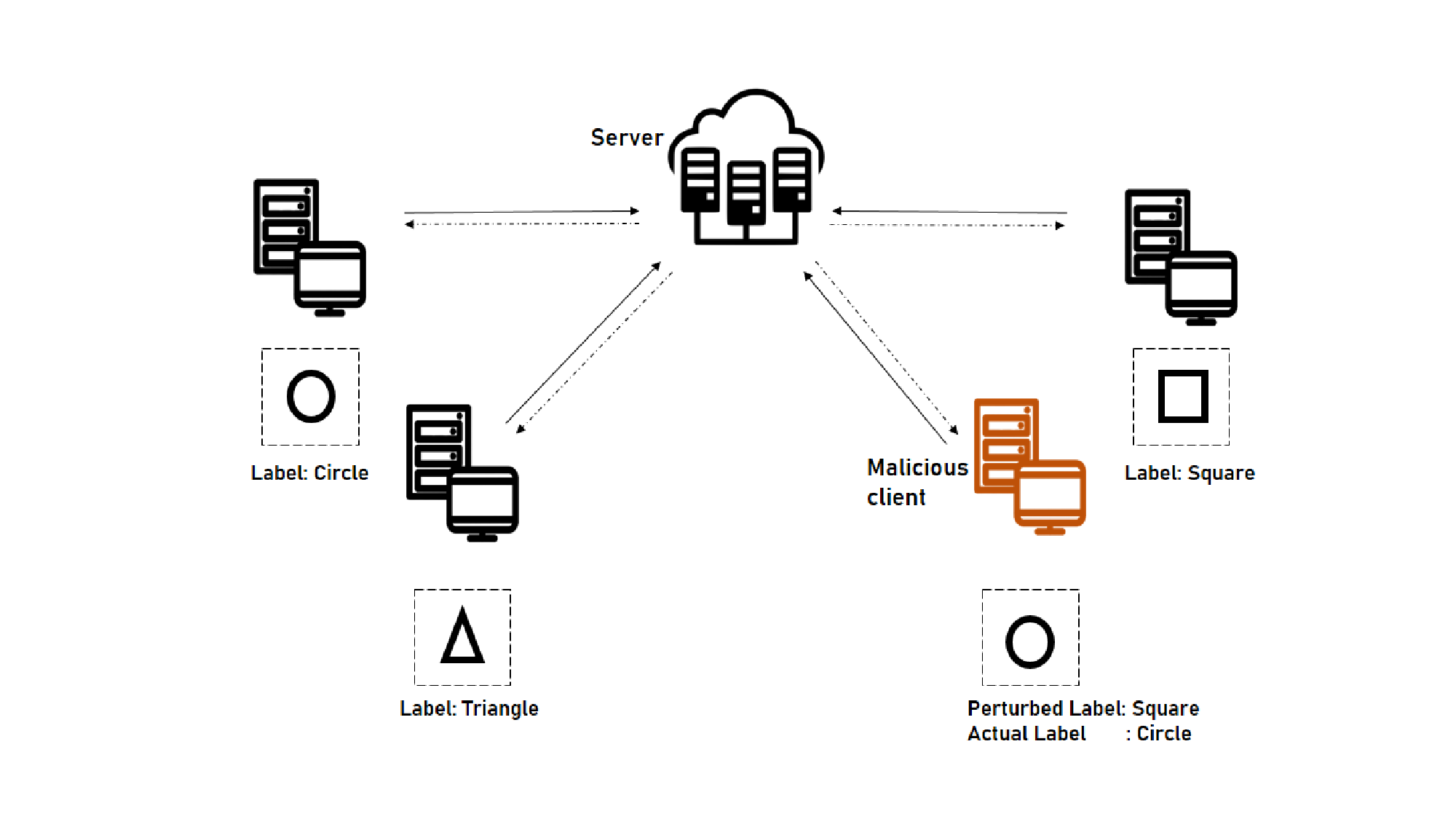}
\centering
\caption{Threat Model}  
\label{fig:fig6}
\end{figure*}

In SFL, participating clients share the smashed data with the server segment of the model which ensures the privacy of the client-training data. Hence, none of the functional components in the framework verifies the quality and security of the training data. Due to the introduction of these vulnerability i.e. privacy and security of client-training data, the server with the split of the global model is now prone to data poisoning attacks from malicious clients in the client group. This paper's threat scenario considers the presence of a subgroup of malicious participants or a percentage of participants who are either malicious or under the control of a malicious adversary. The main objective of the malicious client or the adversary is to poison the training data and compromise the training efficiency. This is carried out by manipulating training by label perturbation. Figure \ref{fig:fig6} illustrates the data poisoning attacks by label flipping in the SFL model. In which one malicious client perturbs the label “circle” of the private training sample to the label “square” and thus infecting the local model.

In this paper, the training data is perturbed by novel targeted, untargeted, and distance-based label-flipping attack algorithms that evade the classifier to produce incorrect results. 
This work considers the following realistic assumptions for the data poisoning attacks:  

\begin{enumerate}

\item This paper considers a realistic scenario where only a percentage of clients are considered malicious or controlled by an external adversary. Given a group of \emph{X} clients, the adversary can take control of \emph{y}\% out of \emph{X} clients. In this research, we evaluate the performance of SFL under varying percentages of malicious clients that match the practicalities of real-time distributed learning scenarios. 

\item The paper assumes  a realistic scenario where each malicious client can only manipulate its private training data.  The adversary or the malicious client cannot influence the aggregation operation of the fed server to produce a global client-side update and does not have access to benign participants’ training data 

\item This paper assumes an honest main server. The assumption of an honest main server is similar to studies that conducted client-side inference attacks \cite{pasquini2021ACM}.

\end{enumerate}
\subsection{Proposed Data Poisoning Attacks strategies}
In this work, the data poisoning attack strategy is implemented by poisoning the labels, that is perturbing the class labels that penultimately causes the trained model to generate incorrect predictions. In SFL, the malicious clients or the adversary trains the client-side model with poisoned training data and transmits the model parameters to the server subsequently influencing the training of the server-side model. Suppose that the given classification task contains \emph{L} classification labels and \emph{l} be the label that is targeted and replaced by label \emph{l'}. Taking the case of this scenario, the attacks introduced in this paper are defined as follows:  

\subsubsection{Targeted Poisoning Attacks}
In the proposed targeted attacks, the adversary selects the labels \emph{l} of source class $S_c$ that the adversary attacks and replaces it with labels \emph{l’} of a target class $T_c$ provided \emph{(l, l’)} $\in$ \emph{L}. Here only the label of $S_c$ is manipulated, and the remaining class labels remain the same. Targeted poisoning attacks aim to reduce the accuracy of the classifier for the targeted source class and the accuracy of remaining non-targeted samples is not affected. Algorithm~\ref{alg:tpa} represents scenario of the targeted poisoning attacks on SFL.  

\begin{algorithm}
\caption{Targeted Poisoning Attack}\label{alg:tpa}
\begin{algorithmic}
\Require $u$ is the $client\_id$ of the malicious client from the given set of \emph{K} clients, \emph{l} is the label of source class $S_c$ and, \emph{l’} is the label of a target class $T_c$. $D^u$ is the dataset of the client $u$ denoted by $\mathrm{\{ x_{i},y_{i} \}}_{i=1}^{m}$, where $x_i$ represents the inputs and,$y_i$ denotes the labels.
\While{local epoch $\emph{e} \neq E$}
\If{the client $u$ is malicious}
    \State iterate through $x_i$ and $y_i$ in $D^u$
    \If{$y_i = = \emph{l}$}
        \State $y_i = \emph{l'}$
    
    \Else
        \State $y_i = \emph{l}$
    
    \EndIf
\EndIf
\EndWhile
\end{algorithmic}
\end{algorithm}

\begin{algorithm}[hbt!]
\caption{Untargeted Poisoning Attack}\label{alg:utpa}
\begin{algorithmic}
\Require $u$ is the client\_id of the malicious client from the given set of \emph{K} clients.$D^u$ is the dataset of the client $u$ denoted by $\mathrm{\{ x_{i},y_{i} \}}_{i=1}^{m}$, where $x_i$ represents the inputs and,$y_i$ denotes the labels.
\While{local epoch $\emph{e} \neq E$}
\If{the client $u$ is malicious}
    \For{\texttt{$y_i$ in $u$}}
        \State \texttt{$y_i = \emph{l}$}
        \State \texttt{where \emph{l} $\in$ \emph{L} is a random label }
    \EndFor
\EndIf
\EndWhile
\end{algorithmic}
\end{algorithm}

\subsubsection{Untargeted Poisoning Attacks}
Proposed untargeted attacks do not target the label of a specific source class, instead, randomly flip a selected set of labels \emph{l} with \emph{l’} where \emph{(l, l’)} $\in$ \emph{L}. Untargeted attacks also flip all class labels to one random class label which drastically reduces the accuracy of the classifier. Untargeted attacks initiated by a set of malicious participants have a greater impact compared to targeted attacks due to the iterative submission of malicious parameters to the server. Algorithm~\ref{alg:utpa} depicts the untargeted poisoning attack. Untargeted attack attempts to degrade the performance of the classifier on the whole rather than the accuracy of a specific class.

\subsubsection{Distance-based Poisoning Attacks}
\begin{algorithm}[hbt!]
\caption{Distance-based Poisoning Attack}\label{alg:dbpa}
\begin{algorithmic}
\Require $u$ is the client\_id of the malicious client from the given set of \emph{K} clients. \emph{l} is the label of source class $S_c$. $D^u$ is the dataset of the client $u$ denoted by $\mathrm{\{ x_{i},y_{i} \}}_{i=1}^{m}$, where $x_i$ represents the inputs and,$y_i$ denotes the labels.
\While{local epoch $\emph{e} \neq E$}
\If{the client $u$ is malicious}
    \State Obtain the indices of the inputs having label such that $y_i = = \emph{l}$
    \State Store the obtained indices in a tensor \emph{indices}
    \If{$len\left( \emph{indices} \right) > 0$}
        \State Select the input that has the label \emph{l}
        \State Compute euclidean distance between selected inputs and other inputs
        \State Obtain the indices of inputs with a maximum distance
        \State Update the label with the label of inputs having a maximum distance
        \State return labels
    \EndIf
\EndIf
\EndWhile
\end{algorithmic}
\end{algorithm}
In the proposed distance-based poisoning attacks, the adversary optimizes and improves the efficiency of targeted attacks by careful selection of target class  $T_c$, and the label of the selected $T_c$ is used to replace the labels of source class $S_c$. To implement a distance-based attack the adversary initially selects a source class $S_c$ with the label \emph{l} and calculates the Euclidean distance between samples of $S_c$ and other training samples where \emph{l} and \emph{l’} are not equal. The Euclidean distance calculates the distance between input samples, provided the input samples are real-valued vectors. 

Further, the training sample with maximum distance is selected and its corresponding class is chosen as the target class $T_c$. The label of this selected target class is used to poison the training samples of $S_c$. The usage of maximum distance is to increase the success rate of the attack. When the source class label is replaced with the label of the sample having a maximum distance, the impact of a poisoning attack is increased.

In a multi-class classification problem, the adversary can select different source class $S_c$ in different trials and initiate attacks by computing the Euclidean distance and identifying target class $T_c$ with maximum. Algorithm~\ref{alg:dbpa} represents distance-based data poisoning attacks that enhance the risk of the ML model since they have a higher impact than targeted poisoning attacks. This type of attack represents a potential threat to SFL in real-world applications.

\section{Implementation} 
\label{cha:implementation}
This section discusses the implementation details of the proposed attack methods.  We describe the datasets involved used  for the research, the model architecture used for training, the experiment setup for SFL, and data poisoning attacks. 
\subsection{Dataset}
We test the proposed methodology using two differnt case studies as mentioned below:

\subsubsection{Case study 1 - Automatic handwritten digit recognition (MNIST Dataset)}
The MNIST dataset is a benchmark dataset for ML and DL classifiers made up of handwritten digits. The dataset consists of 60,000 grayscale images for training and 10,000 grayscale images for testing where the images belong to 10 different classes labelled as ‘0’-'9' Each image is of size 28x28 pixels or 784 features in total \cite{lecun1995mnist}. 

\subsubsection{Case Study 2 - ECG signal classification (ECG Dataset)}
Automatic classification of ECG signals to detect arrhythmia types rules out the need for manual signal analysis by physicians and enables easy monitoring of heart conditions. For this application, the MIT-BIH Arrhythmia dataset that consist of  of ECG signals to classify ECG signals for arrhythmia heartbeat types is used \cite{moody2001ieee}. This standard database contains 48 records, where each record has ECG signals obtained from two separate channels. Each record lasts 30 minutes selected from 24 hours.  Following the analysis of ECG signal processing (\cite{kiranyaz2015ieee}; \cite{de2018robust}; \cite{kuila2020springer}), in this study, 26,490 samples were gathered, and samples also complied with the classification criteria as defined by Association for the Advancement of Medical Instrumentation (AAMI) \cite{association1998ANSI/AAMI}. The collected samples represent 5 different classes of heartbeat types provided in Table \ref{tab:AHT}. Of the total samples half of them are selected randomly to train the model and the remaining are used for the testing process.

\begin{table}[hbt!]
\centering
\caption{Arrhythmia Heartbeat Types}
\label{tab:AHT}

\begin{tabular}{@{}ll@{}}
\toprule
\textbf{Class Label}     & \textbf{Type of Heartbeat} \\ \midrule
N          & Normal beat \\ 
L          & Left bundle branch block \\ 
R          & Right bundle branch block \\ 
A          & Atrial premature contraction \\ 
V          & Ventricular premature contraction\\ 
\bottomrule
\end{tabular}
\end{table}

\subsection{Model Architecture}

We use a 1-Dimensional fully connected dense neural network for the MNIST dataset and a convolutional neural network (1D-CNN) for the ECG dataset. Table~\ref{tab:MD} presents the details of the model architectures. It consists of four convolutional layers with a ReLU activation function, two max-pooling layers, two fully connected dense layers, and a SoftMax activation function that classifies the outputs into one of five categories of arrhythmia heartbeat types.

To train the MNIST dataset, a deep feed-forward network is employed which contains an input layer and 10 dense layers. The final classification layer has a ReLU activation function associated with it to classify the input sample to one of 10 classes in the dataset. 
 
The input layer in the deep feed-forward network is similar to the input layer in 1D-CNN, it receives the input sample from the dataset. The size of the input received in this work for the MNIST dataset is 784 as the images are of size 28x28 pixels.

\begin{table}[hbt!]
\centering
\caption{Models and Datasets}
\label{tab:MD}

\begin{tabular}{@{}llll@{}}
\toprule
\textbf{Dataset} & \textbf{Model}  & \textbf{No: of Labels}  & \textbf{Size of Input}\\ \midrule
ECG dataset & \begin{tabular}[c]{@{}l@{}}1D-CNN\\ 4 Convolutional Layers\\ 2 Dense Layers\end{tabular} & 5 & 124\\ 
MNIST dataset  & \begin{tabular}[c]{@{}l@{}}Feed-forward network\\ 10 dense layers\end{tabular}  & 10 & 784 \\ 
\bottomrule
\end{tabular}
\end{table}

\subsection{SFL Setup}  

For the MNIST dataset, the SFL scenario is defined to have one server and ten clients. The 60,000 training images of the MNIST dataset are partitioned equally among ten clients, where each client has 5000 training records and 1000 validation and testing records. The remaining 10,000 test images are unseen and used for evaluation purposes. The total number of training epochs was finalized as 40 after observing the model convergence rate for different epoch values. 

The experimental setup of SFL for the ECG dataset has one server and five clients. Each client receives distinct and equal batches of data from the train set. The data in the test set is excluded from the training data and it is used for model performance evaluation. The total training epochs are set to 50 as the model convergence is observed in fewer rounds than 50 training epochs.

\subsection{Data Poisoning Attack Setup}

In this paper, to introduce data poisoning attacks only \emph{y}\% of \emph{X} clients are assigned malicious or controlled by an external adversary. The proposed targeted, untargeted, and distance-based attacks were proposed with different percentages of malicious clients for both datasets. For untargeted attacks, all the labels of malicious clients were manipulated and replaced with a class label that has the highest test accuracy in the SFL system. 

In the case of targeted and distance-based attacks, the selection of source class $S_c$ depends on the success of the poisoning attack. In the SFL setting, compromised clients have access to global client-side model updates from the fed server. The malicious client can initiate a poisoning attack for different source classes in a multi-class classification problem and evaluate the impact of the attack that degrades the performance of the classifier. 

In the proposed targeted poisoning attack, the source class $S_c$ is selected as the class that has the highest percentage of correctly identified samples by the classifier. By manipulating the labels of that class with the target class $T_c$ that has the second highest percentage of correctly classified samples. The source class $S_c$ is chosen for distance-based poisoning attacks in an analogous way to targeted attacks. Euclidean distance is computed between inputs that have the label as source class $S_c$ and other inputs. After measuring the distance, the input that has the label as source class $S_c$ is replaced with the label of the input that has a maximum distance. 

In order to increase the impact of the attack, experiments were carried out with different model splits between the client and the server. In the 1D-CNN model for the ECG dataset, the model was split in two positions. At first, the model was split at the second convolutional layer forming two layers for the client segment and four layers for the server segment. Secondly, the model splits at the third convolutional layer forming three layers for the client and three layers for the server segment. The first and the second model splits are called ECGv1 and ECGv2 respectively. Similarly, the deep feed-forward network for the MNIST dataset was also split at two positions, at the second dense layer termed MNISTv1 and at the fourth dense layer referred to as MNISTv2. In the former split, the first two layers form the client-side model, and the remaining eight dense layers belong to the server segment. In the latter split, there will be four layers on the client-side model and six layers on the server-side model. 

\begin{table*}[]
\centering
\caption{Accuracy drop under different attack methods}
\label{tab:acuuracy_comparison}
\begin{tabular}{|cc|cc|cc|cc|}
\hline
\multicolumn{2}{|c|}{}                                    & \multicolumn{2}{c|}{Untargetted attack}      & \multicolumn{2}{c|}{Targetted attack}        & \multicolumn{2}{c|}{\begin{tabular}[c]{@{}c@{}}Distance-based attack\end{tabular}} \\ \hline
\multicolumn{1}{|c|}{}                         & Percentage of malicious clients & \multicolumn{1}{c|}{$A$}     & $A_{d}$ & \multicolumn{1}{c|}{$A$}     & $A_{d}$ & \multicolumn{1}{c|}{$A$}                         & $A_{d}$                     \\ \hline
\multicolumn{1}{|c|}{\multirow{3}{*}{MNISTv1}} & 0        & \multicolumn{1}{c|}{96.46} & 0       & \multicolumn{1}{c|}{96.46} & 0       & \multicolumn{1}{c|}{96.46}                     & 0                            \\ \cline{2-8} 
\multicolumn{1}{|c|}{}                         & 20       & \multicolumn{1}{c|}{95.92} & 0.56     & \multicolumn{1}{c|}{96.13} & 0.34     & \multicolumn{1}{c|}{06.09}                     & 0.38                         \\ \cline{2-8} 
\multicolumn{1}{|c|}{}                         & 40       & \multicolumn{1}{c|}{89.86} & 7.08     & \multicolumn{1}{c|}{91.35} & 5.30     & \multicolumn{1}{c|}{90.77}                     & 5.89                         \\ \hline
\multicolumn{1}{|c|}{\multirow{3}{*}{MNISTv2}} & 0        & \multicolumn{1}{c|}{96.54} & 0        & \multicolumn{1}{c|}{96.54} & 0       & \multicolumn{1}{c|}{96.54}                     & 0                          \\ \cline{2-8} 
\multicolumn{1}{|c|}{}                         & 20       & \multicolumn{1}{c|}{95.54} & 0.62     & \multicolumn{1}{c|}{94.89} & 1.71     & \multicolumn{1}{c|}{95.37}                     & 1.21                         \\ \cline{2-8} 
\multicolumn{1}{|c|}{}                         & 40       & \multicolumn{1}{c|}{86.06} & 11.48    & \multicolumn{1}{c|}{90.58} & 6.17     & \multicolumn{1}{c|}{88.56}                     & 8.26                         \\ \hline
\multicolumn{1}{|c|}{\multirow{3}{*}{ECGV1}}   & 0        & \multicolumn{1}{c|}{88.87} & 0     & \multicolumn{1}{c|}{88.87} & 0        & \multicolumn{1}{c|}{88.87}                     & 0                           \\ \cline{2-8} 
\multicolumn{1}{|c|}{}                         & 20       & \multicolumn{1}{c|}{86.99} & 2.12     & \multicolumn{1}{c|}{88.23} & 0.72     & \multicolumn{1}{c|}{87.99}                     & 0.99                         \\ \cline{2-8} 
\multicolumn{1}{|c|}{}                         & 40       & \multicolumn{1}{c|}{33.87} & 61.89    & \multicolumn{1}{c|}{83.19} & 6.39     & \multicolumn{1}{c|}{76.67}                     & 11.48                        \\ \hline
\multicolumn{1}{|c|}{\multirow{3}{*}{ECGv2}}   & 0        & \multicolumn{1}{c|}{88.89} & 0       & \multicolumn{1}{c|}{88.89} & 0       & \multicolumn{1}{c|}{88.89}                     & 0                           \\ \cline{2-8} 
\multicolumn{1}{|c|}{}                         & 20       & \multicolumn{1}{c|}{75}    & 15.62    & \multicolumn{1}{c|}{87.42} & 1.65     & \multicolumn{1}{c|}{79.77}                     & 10.26                        \\ \cline{2-8} 
\multicolumn{1}{|c|}{}                         & 40       & \multicolumn{1}{c|}{26.50} & 71.31    & \multicolumn{1}{c|}{82.62} & 7.05     & \multicolumn{1}{c|}{75.46}                     & 15.11                        \\ \hline
\end{tabular}
\end{table*}

\section{Results and discussion}
\label{cha:results}
This section examines the effect of data poisoning attacks on MNIST and ECG datasets and the impact of varying the cut-layers.

\subsection{Effects of Data Poisoning Attacks}

This section describes the results of data poisoning attacks on two independent case studies. The effects of targeted, untargeted, and novel distance-based poisoning attacks were examined for each of them.

Table~\ref{tab:acuuracy_comparison} presents the accuracy and accuracy drop ($A_{d}$) in percentage for the two case studies and under different percentages of malicious clients. As seen in the Table, the model's accuracy is greatly reduced due to the untargeted poisoning attack. In the presence of a maximum number of malicious clients, the value of the accuracy drops down to 33.87\% from 88.87\% which results in a 61.89\% drop in accuracy for ECGv1. For ECGv2, the success of the attack is even higher resulting in 71.31\% depletion in accuracy. In ECGv2, it is observable that a small percentage of malicious clients can drastically reduce the accuracy of the model. Thus, the accuracy for MNISTv1 decreased from 96.46\% to 89.86\%. For MNISTv2, a bigger variance in accuracy is seen. When there are 20\% malicious clients present, accuracy falls to 86.06\%.

The success of targeted attacks is low as compared to untargeted attacks. This is due to minimal perturbation in the training samples. The training data of malicious clients contain less corrupt data compared to untargeted scenarios, thereby causing the accuracy to drop by not more than 7\% in either of the split versions.

The accuracy after distance-based attacks is worse than targeted attacks, causing accuracy to drop up to 11.48\% in ECGv1 and 15.11\% in ECGv2. However, the overall accuracy depletion is more for distance-based attacks compared to targeted attacks.   Similar to distance-based attacks induced in the ECG dataset, here the adversary targets a specific class. By manipulating class labels with distance measures, the maximum drop in accuracy is 5.89\% in MNISTv1.  In MNISTv2 the value of $A_{d}$ is 8.26\%.

We also compare the vales of the precision ($P$), Recall ($R$), and F-score ($F$) as provided in the tables~\ref{tab:performance_mnistv1}-\ref{tab:performance_ecgv2}. The table shows the metrics values for different classes. As expected, the metrics change a lot due to the different types of proposed attacks. For example, the precision for the ECGv1 model decreases from 40\% to 10\% for category 1. It should be noted that ECG classification is greatly impacted by the attacks as compared to the MNIST classification data. For example, F-score is only 1\% for category 2 in the ECGv2 model. 

It should be noted that an attacker can adopt various strategies to affect the performance of the model according to their choice and based on their motive.
An attacker can perform untargeted attacks that affect the overall performance of the splitFed-based systems and affect its reliability. Thus, the global model is not able to achieve good performance for any of the classes. The attacker could employ targeted attacks thus affecting the performance of only a specific class rather than the whole global model.  Thus, although it will have higher accuracy, it will induce unfairness in the system as the global model would tend to predict only a specific class.  Similarly, the distance-based attack is a compromise between the two such that it depletes the accuracy as well as impacts specific classes thus inducing unfairness in the system.

\begin{table*}[]
\centering
\caption{Precision, Recall and F-score values for MNISTv1 model}
\label{tab:performance_mnistv1}
\begin{tabular}{|c|ccc|ccc|ccc|ccc|}
\hline
\multicolumn{1}{|l|}{\multirow{2}{*}{Category}} & \multicolumn{3}{l|}{No attack}                                                 & \multicolumn{3}{c|}{Untargetted}                             & \multicolumn{3}{c|}{Targetted}                                                 & \multicolumn{3}{c|}{\begin{tabular}[c]{@{}c@{}}Distance\\ based\end{tabular}}  \\ \cline{2-13} 
\multicolumn{1}{|l|}{}                          & \multicolumn{1}{l|}{P}    & \multicolumn{1}{l|}{R}    & \multicolumn{1}{l|}{F} & \multicolumn{1}{c|}{P}    & \multicolumn{1}{c|}{R}    & F    & \multicolumn{1}{c|}{P}    & \multicolumn{1}{c|}{R}    & \multicolumn{1}{l|}{F} & \multicolumn{1}{c|}{P}    & \multicolumn{1}{c|}{R}    & \multicolumn{1}{l|}{F} \\ \hline
0                                               & \multicolumn{1}{c|}{0.98} & \multicolumn{1}{c|}{0.97} & 0.98                   & \multicolumn{1}{c|}{0.88} & \multicolumn{1}{c|}{0.90} & 0.89 & \multicolumn{1}{c|}{0.98} & \multicolumn{1}{c|}{0.94} & 0.96                   & \multicolumn{1}{c|}{0.98} & \multicolumn{1}{c|}{0.89} & 0.94                   \\ \hline
1                                               & \multicolumn{1}{c|}{0.98} & \multicolumn{1}{c|}{0.99} & 0.98                   & \multicolumn{1}{c|}{0.83} & \multicolumn{1}{c|}{0.98} & 0.90 & \multicolumn{1}{c|}{0.61} & \multicolumn{1}{c|}{0.99} & 0.76                   & \multicolumn{1}{c|}{0.43} & \multicolumn{1}{c|}{0.88} & 0.58                   \\ \hline
2                                               & \multicolumn{1}{c|}{0.94} & \multicolumn{1}{c|}{0.98} & 0.96                   & \multicolumn{1}{c|}{0.85} & \multicolumn{1}{c|}{0.88} & 0.86 & \multicolumn{1}{c|}{0.93} & \multicolumn{1}{c|}{0.92} & 0.92                   & \multicolumn{1}{c|}{0.94} & \multicolumn{1}{c|}{0.93} & 0.93                   \\ \hline
3                                               & \multicolumn{1}{c|}{0.94} & \multicolumn{1}{c|}{0.96} & 0.95                   & \multicolumn{1}{c|}{0.80} & \multicolumn{1}{c|}{0.89} & 0.84 & \multicolumn{1}{c|}{0.89} & \multicolumn{1}{c|}{0.90} & 0.91                   & \multicolumn{1}{c|}{0.85} & \multicolumn{1}{c|}{0.93} & 0.89                   \\ \hline
4                                               & \multicolumn{1}{c|}{0.98} & \multicolumn{1}{c|}{0.96} & 0.97                   & \multicolumn{1}{c|}{0.91} & \multicolumn{1}{c|}{0.89} & 0.90 & \multicolumn{1}{c|}{0.93} & \multicolumn{1}{c|}{0.93} & 0.93                   & \multicolumn{1}{c|}{0.95} & \multicolumn{1}{c|}{0.94} & 0.94                   \\ \hline
5                                               & \multicolumn{1}{c|}{0.95} & \multicolumn{1}{c|}{0.94} & 0.95                   & \multicolumn{1}{c|}{0.91} & \multicolumn{1}{c|}{0.46} & 0.61 & \multicolumn{1}{c|}{0.95} & \multicolumn{1}{c|}{0.82} & 0.88                   & \multicolumn{1}{c|}{0.98} & \multicolumn{1}{c|}{0.62} & 0.76                   \\ \hline
6                                               & \multicolumn{1}{c|}{0.98} & \multicolumn{1}{c|}{1}    & 0.99                   & \multicolumn{1}{c|}{0.92} & \multicolumn{1}{c|}{0.94} & 0.93 & \multicolumn{1}{c|}{0.95} & \multicolumn{1}{c|}{0.93} & 0.94                   & \multicolumn{1}{c|}{0.98} & \multicolumn{1}{c|}{0.95} & 0.96                   \\ \hline
7                                               & \multicolumn{1}{c|}{0.99} & \multicolumn{1}{c|}{0.98} & 0.99                   & \multicolumn{1}{c|}{0.93} & \multicolumn{1}{c|}{0.91} & 0.92 & \multicolumn{1}{c|}{0.93} & \multicolumn{1}{c|}{0.94} & 0.93                   & \multicolumn{1}{c|}{0.97} & \multicolumn{1}{c|}{0.94} & 0.95                   \\ \hline
8                                               & \multicolumn{1}{c|}{0.98} & \multicolumn{1}{c|}{0.92} & 0.95                   & \multicolumn{1}{c|}{0.74} & \multicolumn{1}{c|}{0.67} & 0.70 & \multicolumn{1}{c|}{0.84} & \multicolumn{1}{c|}{0.86} & 0.84                   & \multicolumn{1}{c|}{0.75} & \multicolumn{1}{c|}{0.87} & 0.82                   \\ \hline
9                                               & \multicolumn{1}{c|}{0.94} & \multicolumn{1}{c|}{0.95} & 0.95                   & \multicolumn{1}{c|}{0.84} & \multicolumn{1}{c|}{0.81} & 0.82 & \multicolumn{1}{c|}{0.96} & \multicolumn{1}{c|}{0.86} & 0.91                   & \multicolumn{1}{c|}{0.94} & \multicolumn{1}{c|}{0.87} & 0.90                   \\ \hline
\end{tabular}
\end{table*}

\begin{table*}[]
\centering
\caption{Precision, Recall and F-score values for MNISTv2 model}
\label{tab:performance_mnistv2}
\begin{tabular}{|c|ccc|ccc|ccc|ccc|}
\hline
\multicolumn{1}{|l|}{\multirow{2}{*}{Category}} & \multicolumn{3}{l|}{No attack}                                                 & \multicolumn{3}{c|}{Untargetted}                             & \multicolumn{3}{c|}{Targetted}                                                 & \multicolumn{3}{c|}{\begin{tabular}[c]{@{}c@{}}Distance\\ based\end{tabular}}  \\ \cline{2-13} 
\multicolumn{1}{|l|}{}                          & \multicolumn{1}{l|}{P}    & \multicolumn{1}{l|}{R}    & \multicolumn{1}{l|}{F} & \multicolumn{1}{c|}{P}    & \multicolumn{1}{c|}{R}    & F    & \multicolumn{1}{c|}{P}    & \multicolumn{1}{c|}{R}    & \multicolumn{1}{l|}{F} & \multicolumn{1}{c|}{P}    & \multicolumn{1}{c|}{R}    & \multicolumn{1}{l|}{F} \\ \hline
0                                               & \multicolumn{1}{c|}{0.98} & \multicolumn{1}{c|}{0.97} & 0.98                   & \multicolumn{1}{c|}{0.88} & \multicolumn{1}{c|}{0.90} & 0.89 & \multicolumn{1}{c|}{0.98} & \multicolumn{1}{c|}{0.94} & 0.96                   & \multicolumn{1}{c|}{0.99} & \multicolumn{1}{c|}{0.89} & 0.94                   \\ \hline
1                                               & \multicolumn{1}{c|}{0.98} & \multicolumn{1}{c|}{0.99} & 0.98                   & \multicolumn{1}{c|}{0.85} & \multicolumn{1}{c|}{0.98} & 0.91 & \multicolumn{1}{c|}{0.63} & \multicolumn{1}{c|}{0.97} & 0.76                   & \multicolumn{1}{c|}{0.40} & \multicolumn{1}{c|}{0.85} & 0.55                   \\ \hline
2                                               & \multicolumn{1}{c|}{0.94} & \multicolumn{1}{c|}{0.98} & 0.96                   & \multicolumn{1}{c|}{0.87} & \multicolumn{1}{c|}{0.88} & 0.87 & \multicolumn{1}{c|}{0.93} & \multicolumn{1}{c|}{0.92} & 0.92                   & \multicolumn{1}{c|}{0.94} & \multicolumn{1}{c|}{0.93} & 0.93                   \\ \hline
3                                               & \multicolumn{1}{c|}{0.94} & \multicolumn{1}{c|}{0.96} & 0.95                   & \multicolumn{1}{c|}{0.80} & \multicolumn{1}{c|}{0.89} & 0.84 & \multicolumn{1}{c|}{0.89} & \multicolumn{1}{c|}{0.93} & 0.91                   & \multicolumn{1}{c|}{0.85} & \multicolumn{1}{c|}{0.93} & 0.89                   \\ \hline
4                                               & \multicolumn{1}{c|}{0.98} & \multicolumn{1}{c|}{0.96} & 0.97                   & \multicolumn{1}{c|}{0.91} & \multicolumn{1}{c|}{0.89} & 0.90 & \multicolumn{1}{c|}{0.93} & \multicolumn{1}{c|}{0.93} & 0.93                   & \multicolumn{1}{c|}{0.95} & \multicolumn{1}{c|}{0.94} & 0.94                   \\ \hline
5                                               & \multicolumn{1}{c|}{0.95} & \multicolumn{1}{c|}{0.94} & 0.95                   & \multicolumn{1}{c|}{0.98} & \multicolumn{1}{c|}{0.46} & 0.62 & \multicolumn{1}{c|}{0.95} & \multicolumn{1}{c|}{0.82} & 0.88                   & \multicolumn{1}{c|}{0.98} & \multicolumn{1}{c|}{0.62} & 0.76                   \\ \hline
6                                               & \multicolumn{1}{c|}{0.98} & \multicolumn{1}{c|}{1}    & 0.99                   & \multicolumn{1}{c|}{0.92} & \multicolumn{1}{c|}{0.94} & 0.93 & \multicolumn{1}{c|}{0.96} & \multicolumn{1}{c|}{0.93} & 0.94                   & \multicolumn{1}{c|}{0.98} & \multicolumn{1}{c|}{0.95} & 0.96                   \\ \hline
7                                               & \multicolumn{1}{c|}{0.99} & \multicolumn{1}{c|}{0.98} & 0.99                   & \multicolumn{1}{c|}{0.94} & \multicolumn{1}{c|}{0.91} & 0.92 & \multicolumn{1}{c|}{0.93} & \multicolumn{1}{c|}{0.94} & 0.93                   & \multicolumn{1}{c|}{0.97} & \multicolumn{1}{c|}{0.94} & 0.95                   \\ \hline
8                                               & \multicolumn{1}{c|}{0.98} & \multicolumn{1}{c|}{0.92} & 0.95                   & \multicolumn{1}{c|}{0.73} & \multicolumn{1}{c|}{0.67} & 0.70 & \multicolumn{1}{c|}{0.87} & \multicolumn{1}{c|}{0.85} & 0.86                   & \multicolumn{1}{c|}{0.76} & \multicolumn{1}{c|}{0.89} & 0.82                   \\ \hline
9                                               & \multicolumn{1}{c|}{0.94} & \multicolumn{1}{c|}{0.95} & 0.95                   & \multicolumn{1}{c|}{0.84} & \multicolumn{1}{c|}{0.81} & 0.82 & \multicolumn{1}{c|}{0.96} & \multicolumn{1}{c|}{0.86} & 0.91                   & \multicolumn{1}{c|}{0.94} & \multicolumn{1}{c|}{0.87} & 0.90                   \\ \hline
\end{tabular}
\end{table*}

\begin{table*}[]
\centering
\caption{Precision, Recall and F-score values for ECGv1 model}
\label{tab:performance_ECGv1}
\begin{tabular}{|c|ccc|ccc|ccc|ccc|}
\hline
\multicolumn{1}{|l|}{\multirow{2}{*}{Category}} & \multicolumn{3}{l|}{No attack}                                                 & \multicolumn{3}{c|}{Untargetted}                             & \multicolumn{3}{c|}{Targetted}                                                 & \multicolumn{3}{c|}{\begin{tabular}[c]{@{}c@{}}Distance\\ based\end{tabular}}  \\ \cline{2-13} 
\multicolumn{1}{|l|}{}                          & \multicolumn{1}{l|}{P}    & \multicolumn{1}{l|}{R}    & \multicolumn{1}{l|}{F} & \multicolumn{1}{c|}{P}    & \multicolumn{1}{c|}{R}    & F    & \multicolumn{1}{c|}{P}    & \multicolumn{1}{c|}{R}    & \multicolumn{1}{l|}{F} & \multicolumn{1}{c|}{P}    & \multicolumn{1}{c|}{R}    & \multicolumn{1}{l|}{F} \\ \hline
0                                               & \multicolumn{1}{c|}{0.80} & \multicolumn{1}{c|}{0.96} & 0.87                   & \multicolumn{1}{c|}{0.80} & \multicolumn{1}{c|}{015}  & 0.25 & \multicolumn{1}{c|}{0.55} & \multicolumn{1}{c|}{0.60} & 0.57                   & \multicolumn{1}{c|}{0.78} & \multicolumn{1}{c|}{0.77} & 0.78                   \\ \hline
1                                               & \multicolumn{1}{c|}{0.86} & \multicolumn{1}{c|}{0.99} & 0.92                   & \multicolumn{1}{c|}{0.75} & \multicolumn{1}{c|}{010}  & 017  & \multicolumn{1}{c|}{0.27} & \multicolumn{1}{c|}{0.15} & 0.18                   & \multicolumn{1}{c|}{0.23} & \multicolumn{1}{c|}{0.24} & 0.23                   \\ \hline
2                                               & \multicolumn{1}{c|}{0.79} & \multicolumn{1}{c|}{0.78} & 0.79                   & \multicolumn{1}{c|}{0.70} & \multicolumn{1}{c|}{0.05} & 0.09 & \multicolumn{1}{c|}{0.50} & \multicolumn{1}{c|}{0.40} & 0.44                   & \multicolumn{1}{c|}{0.50} & \multicolumn{1}{c|}{0.51} & 0.51                   \\ \hline
3                                               & \multicolumn{1}{c|}{0.40} & \multicolumn{1}{c|}{0.50} & 0.44                   & \multicolumn{1}{c|}{0.10} & \multicolumn{1}{c|}{0.97} & 0.18 & \multicolumn{1}{c|}{0.75} & \multicolumn{1}{c|}{0.78} & 0.77                   & \multicolumn{1}{c|}{0.86} & \multicolumn{1}{c|}{0.76} & 0.81                   \\ \hline
4                                               & \multicolumn{1}{c|}{0.90} & \multicolumn{1}{c|}{0.97} & 0.94                   & \multicolumn{1}{c|}{0.95} & \multicolumn{1}{c|}{0.55} & 0.69 & \multicolumn{1}{c|}{0.85} & \multicolumn{1}{c|}{0.90} & 0.88                   & \multicolumn{1}{c|}{0.86} & \multicolumn{1}{c|}{0.90} & 0.87                   \\ \hline
\end{tabular}
\end{table*}

\begin{table*}[]
\centering
\caption{Precision, Recall and F-score values for ECGv2 model}
\label{tab:performance_ecgv2}
\begin{tabular}{|c|ccc|ccc|ccc|ccc|}
\hline
\multicolumn{1}{|l|}{\multirow{2}{*}{Category}} & \multicolumn{3}{l|}{No attack}                                                 & \multicolumn{3}{c|}{Untargetted}                             & \multicolumn{3}{c|}{Targetted}                                                 & \multicolumn{3}{c|}{\begin{tabular}[c]{@{}c@{}}Distance\\ based\end{tabular}}  \\ \cline{2-13} 
\multicolumn{1}{|l|}{}                          & \multicolumn{1}{l|}{P}    & \multicolumn{1}{l|}{R}    & \multicolumn{1}{l|}{F} & \multicolumn{1}{c|}{P}    & \multicolumn{1}{c|}{R}    & F    & \multicolumn{1}{c|}{P}    & \multicolumn{1}{c|}{R}    & \multicolumn{1}{l|}{F} & \multicolumn{1}{c|}{P}    & \multicolumn{1}{c|}{R}    & \multicolumn{1}{l|}{F} \\ \hline
0                                               & \multicolumn{1}{c|}{0.80} & \multicolumn{1}{c|}{0.96} & 0.87                   & \multicolumn{1}{c|}{0.83} & \multicolumn{1}{c|}{0.14} & 0.23 & \multicolumn{1}{c|}{0.60} & \multicolumn{1}{c|}{0.65} & 0.62                   & \multicolumn{1}{c|}{0.70} & \multicolumn{1}{c|}{0.77} & 0.73                   \\ \hline
1                                               & \multicolumn{1}{c|}{0.86} & \multicolumn{1}{c|}{0.99} & 0.92                   & \multicolumn{1}{c|}{0.73} & \multicolumn{1}{c|}{0.06} & 0.11 & \multicolumn{1}{c|}{0.20} & \multicolumn{1}{c|}{0.10} & 0.13                   & \multicolumn{1}{c|}{0.20} & \multicolumn{1}{c|}{0.22} & 0.21                   \\ \hline
2                                               & \multicolumn{1}{c|}{0.79} & \multicolumn{1}{c|}{0.78} & 0.79                   & \multicolumn{1}{c|}{0.62} & \multicolumn{1}{c|}{0.01} & 0.01 & \multicolumn{1}{c|}{0.55} & \multicolumn{1}{c|}{0.45} & 0.50                   & \multicolumn{1}{c|}{0.52} & \multicolumn{1}{c|}{0.51} & 0.52                   \\ \hline
3                                               & \multicolumn{1}{c|}{0.40} & \multicolumn{1}{c|}{0.50} & 0.44                   & \multicolumn{1}{c|}{0.11} & \multicolumn{1}{c|}{0.98} & 0.20 & \multicolumn{1}{c|}{0.80} & \multicolumn{1}{c|}{0.83} & 0.82                   & \multicolumn{1}{c|}{0.87} & \multicolumn{1}{c|}{0.76} & 0.81                   \\ \hline
4                                               & \multicolumn{1}{c|}{0.90} & \multicolumn{1}{c|}{0.97} & 0.94                   & \multicolumn{1}{c|}{0.96} & \multicolumn{1}{c|}{0.58} & 0.73 & \multicolumn{1}{c|}{0.90} & \multicolumn{1}{c|}{0.95} & 0.92                   & \multicolumn{1}{c|}{0.87} & \multicolumn{1}{c|}{0.89} & 0.88                   \\ \hline
\end{tabular}
\end{table*}
\subsection{Impact of Changing Cut Layers}
\begin{figure}[h]
\includegraphics[width=1\linewidth]{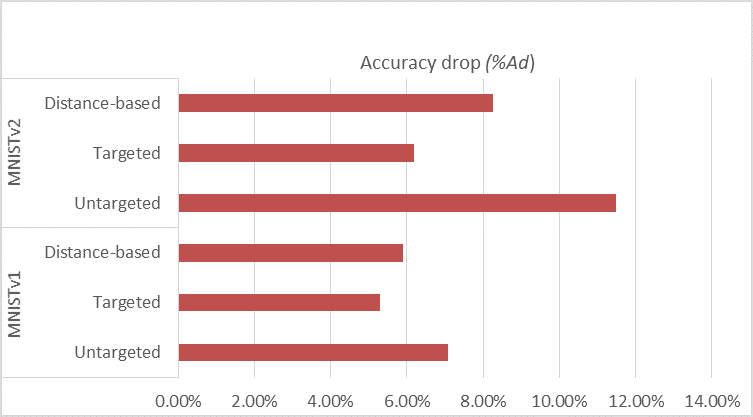}
\centering
\caption{Accuracy drop $A_{d}$ v/s Split Layer in MNIST Dataset }  
\label{fig:fig11}
\end{figure}

\begin{figure}[h]
\includegraphics[width=1\linewidth]{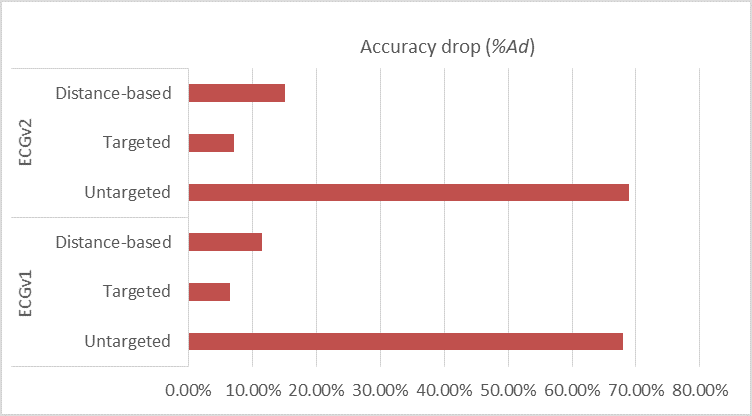}
\centering
\caption{Accuracy drop $A_{d}$ v/s Split Layer in ECG Dataset }  
\label{fig:fig12}
\end{figure}

The layer at which the model is divided between the client and server in the SFL has a serious influence on how effective poisoning attempts are. Attack intensity also varies with different cut layer choices. It is evident from the numerical data in both case studies that the poisoning attack on MNISTv2 and ECGv2 is more effective since these versions produce greater values of $A_{d}$. The reason for this is that there are now more layers in the client segment, giving the adversary greater room to initiate a more powerful and efficient attack. However, with a smaller number of model layers on the client segment, the model's overall accuracy is not greatly affected. Figure \ref{fig:fig11} and \ref{fig:fig12} depicts the relationship between accuracy drop $A_{d}$ and cut layer observed from the experimental results of the two case studies. 

\subsection{Accuracy Depletion with Changing Percentage of Malicious Clients}
The percentage of malicious clients plays a vital role in degrading the model accuracy during data poisoning attacks. Increasing the value of malicious clients in the SFL setting can drastically reduce accuracy. Considering the possibilities of a practical scenario, it is not ideal to have a large number of malicious clients in the SFL system. In this paper, the depletion of accuracy is studied with a varied percentage of malicious clients. The results of the two case studies make it clear that even with 10\% of malicious clients, the accuracy value falls to a certain level. In untargeted attacks, the higher the percentage of malicious clients, the higher the value of accuracy drops. 40\% of malicious clients in ECGv2 causes the accuracy to drop from 88.89\% to 26.50\%. As expected, the results of all three attack strategies clearly conclude that increasing the percentage of malicious clients contributes to the success of data poisoning attacks. 

After the critical analysis of experimental results, untargeted attacks have a significant impact on the classifier results. However, an attacker can still adopt targeted or distance-based attacks to reduce the classifier performance for a specific class. By adopting this strategy, it is possible to initiate attacks that cannot be directly detected while still maintaining better accuracy. This can degrade the classifier performance for one specific class chosen by the adversary.

\section{Conclusions}  
\label{conclusions}
This paper is the initial attempt to investigate the effectiveness of various types of data poisoning attacks against SFL. 
The performance of the attack strategy is evaluated under several factors such as the number of split layers between the client and server and varying percentages of malicious clients in the SFL setting. An important indicator that shows how accuracy decreases with attack intensity is the value of accuracy drop. 
Distance-based data poisoning attacks have higher efficacy than targeted attacks. The highest value of accuracy drop resulted due to distance-based attack is 8.26\% for the MNIST dataset and 15.11\% for the ECG dataset. Furthermore, it can be concluded the SFL is more vulnerable to untargeted attacks which deteriorate the overall performance of the classifier. In addition to that SFL is more susceptible to distance-based data poisoning attacks compared to conventional targeted poisoning attacks. 
It should be noted that an attacker can employ targeted or distance-based attacks to lower the performance of the classifier for a particular class. Thereby, launching attacks that are difficult to detect immediately yet maintains a good overall accuracy. As a result, the performance of the classifier for a selected class is degraded. This research revealed the risk and vulnerability of SFL from the empirical results obtained after inducing data poisoning attacks with malicious clients.


\bibliographystyle{ieeetr}
\bibliography{references}

\begin{thebibliography}{10}

\bibitem{howarth_2023}
J.~Howarth, ``\textit{57+ amazing artificial intelligence statistics (2023)},''
  Feb 2023.
\newblock https://explodingtopics.com/blog/ai-statistics.

\bibitem{shetty2022supervised}
S.~H. Shetty, S.~Shetty, C.~Singh, and A.~Rao, ``Supervised machine learning:
  Algorithms and applications,'' {\em Fundamentals and Methods of Machine and
  Deep Learning: Algorithms, Tools and Applications}, pp.~1--16, 2022.

\bibitem{hu2019ACM}
Y.~Hu, D.~Niu, J.~Yang, and S.~Zhou, ``Fdml: A collaborative machine learning
  framework for distributed features,'' in {\em \textit{Proceedings of the 25th
  ACM SIGKDD International Conference on Knowledge Discovery \& Data Mining}},
  pp.~2232--2240, 2019.

\bibitem{bansal2022acm}
M.~A. Bansal, D.~R. Sharma, and D.~M. Kathuria, ``A systematic review on data
  scarcity problem in deep learning: solution and applications,'' {\em ACM
  Computing Surveys (CSUR)}, vol.~54, no.~10s, pp.~1--29, 2022.

\bibitem{guo2021arxiv}
S.~Guo, X.~Zhang, F.~Yang, T.~Zhang, Y.~Gan, T.~Xiang, and Y.~Liu, ``Robust and
  privacy-preserving collaborative learning: A comprehensive survey,'' {\em
  \textit{arXiv preprint arXiv:2112.10183}}, 2021.

\bibitem{li2020ieee}
T.~Li, A.~K. Sahu, A.~Talwalkar, and V.~Smith, ``Federated learning:
  Challenges, methods, and future directions,'' {\em \textit{IEEE signal
  processing magazine}}, vol.~37, no.~3, pp.~50--60, 2020.

\bibitem{vepakomma2018arxiv}
P.~Vepakomma, O.~Gupta, T.~Swedish, and R.~Raskar, ``Split learning for health:
  Distributed deep learning without sharing raw patient data,'' {\em
  \textit{arXiv preprint arXiv:1812.00564}}, 2018.

\bibitem{gupta2018Elsevier}
O.~Gupta and R.~Raskar, ``Distributed learning of deep neural network over
  multiple agents,'' {\em \textit{Journal of Network and Computer
  Applications}}, vol.~116, pp.~1--8, 2018.

\bibitem{thapa2022AAAI}
C.~Thapa, P.~C.~M. Arachchige, S.~Camtepe, and L.~Sun, ``Splitfed: When
  federated learning meets split learning,'' in {\em \textit{Proceedings of the
  AAAI Conference on Artificial Intelligence}}, vol.~36, pp.~8485--8493, 2022.

\bibitem{fang2020USENIX}
M.~Fang, X.~Cao, J.~Jia, and N.~Z. Gong, ``Local model poisoning attacks to
  byzantine-robust federated learning,'' in {\em \textit{Proceedings of the
  29th USENIX Conference on Security Symposium}}, pp.~1623--1640, 2020.

\bibitem{shejwalkar2022ieee}
V.~Shejwalkar, A.~Houmansadr, P.~Kairouz, and D.~Ramage, ``Back to the drawing
  board: A critical evaluation of poisoning attacks on production federated
  learning,'' in {\em \textit{2022 IEEE Symposium on Security and Privacy
  (SP)}}, pp.~1354--1371, IEEE, 2022.

\bibitem{pasquini2021ACM}
D.~Pasquini, G.~Ateniese, and M.~Bernaschi, ``Unleashing the tiger: Inference
  attacks on split learning,'' in {\em \textit{Proceedings of the 2021 ACM
  SIGSAC Conference on Computer and Communications Security}}, pp.~2113--2129,
  2021.

\bibitem{lin2021arxiv}
J.~Lin, L.~Dang, M.~Rahouti, and K.~Xiong, ``Ml attack models: Adversarial
  attacks and data poisoning attacks,'' {\em \textit{arXiv preprint
  arXiv:2112.02797}}, 2021.

\bibitem{sun2022ieee}
Y.~Sun, H.~Ochiai, and J.~Sakuma, ``Semi-targeted model poisoning attack on
  federated learning via backward error analysis,'' in {\em 2022 International
  Joint Conference on Neural Networks (IJCNN)}, pp.~1--8, IEEE, 2022.

\bibitem{duan2022sensors}
Q.~Duan, S.~Hu, R.~Deng, and Z.~Lu, ``Combined federated and split learning in
  edge computing for ubiquitous intelligence in internet of things:
  State-of-the-art and future directions,'' {\em \textit{Sensors}}, vol.~22,
  no.~16, p.~5983, 2022.

\bibitem{lin2020ACM}
J.~Lin, L.~Xu, Y.~Liu, and X.~Zhang, ``Composite backdoor attack for deep
  neural network by mixing existing benign features,'' in {\em
  \textit{Proceedings of the 2020 ACM SIGSAC Conference on Computer and
  Communications Security}}, pp.~113--131, 2020.

\bibitem{lyu2020arxiv}
L.~Lyu, H.~Yu, and Q.~Yang, ``Threats to federated learning: A survey,'' {\em
  \textit{arXiv preprint arXiv:2003.02133}}, 2020.

\bibitem{lee2021IEEE}
D.~Lee, J.~Lee, H.~Jun, H.~Kim, and S.~Yoo, ``Triad of split learning: Privacy,
  accuracy, and performance,'' in {\em \textit{2021 International Conference on
  Information and Communication Technology Convergence (ICTC)}},
  pp.~1185--1189, IEEE, 2021.

\bibitem{erdougan2022unsplit}
E.~Erdo{\u{g}}an, A.~K{\"u}p{\c{c}}{\"u}, and A.~E. {\c{C}}i{\c{c}}ek,
  ``Unsplit: Data-oblivious model inversion, model stealing, and label
  inference attacks against split learning,'' in {\em \textit{Proceedings of
  the 21st Workshop on Privacy in the Electronic Society}}, pp.~115--124, 2022.

\bibitem{li2021arxiv}
O.~Li, J.~Sun, X.~Yang, W.~Gao, H.~Zhang, J.~Xie, V.~Smith, and C.~Wang,
  ``Label leakage and protection in two-party split learning,'' {\em
  \textit{arXiv preprint arXiv:2102.08504}}.

\bibitem{vepakomma2020ieee}
P.~Vepakomma, A.~Singh, O.~Gupta, and R.~Raskar, ``Nopeek: Information leakage
  reduction to share activations in distributed deep learning,'' in {\em
  \textit{2020 International Conference on Data Mining Workshops (ICDMW)}},
  pp.~933--942, IEEE, 2020.

\bibitem{titcombe2021arxiv}
T.~Titcombe, A.~J. Hall, P.~Papadopoulos, and D.~Romanini, ``Practical defences
  against model inversion attacks for split neural networks,'' {\em
  \textit{arXiv preprint arXiv:2104.05743}}, 2021.

\bibitem{thapa2021springer}
C.~Thapa, M.~A.~P. Chamikara, and S.~A. Camtepe, ``Advancements of federated
  learning towards privacy preservation: from federated learning to split
  learning,'' {\em \textit{Federated Learning Systems: Towards Next-Generation
  AI}}, pp.~79--109, 2021.

\bibitem{GajbhiyeRTIMPR2022}
S.~Gajbhiye, P.~Singh, and S.~Gupta, ``Data poisoning attack by label flipping
  on splitfed learning,'' in {\em International Conference on Recent Trends in
  Image Processing and Pattern Recognition}, pp.~391--405, Springer, 2022.

\bibitem{mcmahan2017ais}
B.~McMahan, E.~Moore, D.~Ramage, S.~Hampson, and B.~A. y~Arcas,
  ``Communication-efficient learning of deep networks from decentralized
  data,'' in {\em \textit{Artificial intelligence and statistics}},
  pp.~1273--1282, PMLR, 2017.

\bibitem{lecun1995mnist}
Y.~LeCun, L.~Jackel, L.~Bottou, A.~Brunot, C.~Cortes, J.~Denker, H.~Drucker,
  I.~Guyon, U.~Muller, E.~Sackinger, {\em et~al.}, ``Comparison of learning
  algorithms for handwritten digit recognition,'' in {\em \textit{International
  conference on artificial neural networks}}, vol.~60, pp.~53--60, Perth,
  Australia, 1995.

\bibitem{moody2001ieee}
G.~B. Moody and R.~G. Mark, ``The impact of the mit-bih arrhythmia database,''
  {\em \textit{IEEE engineering in medicine and biology magazine}}, vol.~20,
  no.~3, pp.~45--50, 2001.

\bibitem{kiranyaz2015ieee}
S.~Kiranyaz, T.~Ince, and M.~Gabbouj, ``Real-time patient-specific ecg
  classification by 1-d convolutional neural networks,'' {\em \textit{IEEE
  Transactions on Biomedical Engineering}}, vol.~63, no.~3, pp.~664--675, 2015.

\bibitem{de2018robust}
V.~H.~C. De~Albuquerque, T.~M. Nunes, D.~R. Pereira, E.~J. d.~S. Luz,
  D.~Menotti, J.~P. Papa, and J.~M.~R. Tavares, ``Robust automated cardiac
  arrhythmia detection in ecg beat signals,'' {\em \textit{Neural Computing and
  Applications}}, vol.~29, pp.~679--693, 2018.

\bibitem{kuila2020springer}
S.~Kuila, N.~Dhanda, and S.~Joardar, ``Feature extraction and classification of
  mit-bih arrhythmia database,'' in {\em \textit{Proceedings of the 2nd
  International Conference on Communication, Devices and Computing: ICCDC
  2019}}, pp.~417--427, Springer, 2020.

\bibitem{association1998ANSI/AAMI}
N.~Association for the Advancement~of Medical~Instrumentation {\em et~al.},
  ``Testing and reporting performance results of cardiac rhythm and st segment
  measurement algorithms,'' {\em \textit{ANSI/AAMI EC38}}, vol.~1998, p.~46,
  1998.

\end{thebibliography}
\end{document}